\documentclass[conference]{IEEEtran}
\IEEEoverridecommandlockouts
\usepackage{cite}
\usepackage{amsmath,amssymb,amsfonts}
\usepackage{algorithmic}
\usepackage{graphicx}
\usepackage{textcomp}
\usepackage{xcolor}
\usepackage{multirow}
\usepackage{booktabs}
\usepackage{float}
\usepackage[colorlinks,linkcolor=blue]{hyperref}
\makeatletter
\let\NAT@parse\undefined
\makeatother
\usepackage{hyperref}  

\def\BibTeX{{\rm B\kern-.05em{\sc i\kern-.025em b}\kern-.08em
    T\kern-.1667em\lower.7ex\hbox{E}\kern-.125emX}}
\begin{document}

\title{STSA: Spatial-Temporal Semantic Alignment for Visual Dubbing\\

\thanks{This work was partially supported by NSFC grant 12425113; in part by the Natural Science Foundation of Jiangsu Province under Grant BK20240462; in part by the China Postdoctoral Science Foundation under Grant 2023M733427 and Grant 2023TQ0349; and the Jiangsu Funding Program for Excellent Postdoctoral Talent.}
}


\author{\IEEEauthorblockN{\textit{Zijun Ding$^{1,2}$, Mingdie Xiong$^{3}$, Congcong Zhu$^{1,2, \ast}$, Jingrun Chen$^{1,2}$} \thanks{$^{\ast}$Corresponding author.} }
\IEEEauthorblockA{
\text{$^{1}$Key Laboratory of the Ministry of Education for Mathematical Foundations and Applications of Digital Technology, } \\
\text{University of Science and Technology of China, Hefei, China}\\
\text{$^{2}$Suzhou Institute for Advanced Research, USTC, Suzhou, China} 
\text{$^{3}$Harbin Engineering University, Harbin, China} \\
\text{zijunding@mail.ustc.edu.cn, xiongmingdie@hrbeu.edu.cn, \{cczly, jingrunchen\}@ustc.edu.cn}\\
}
}

\maketitle

\begin{abstract}
Existing audio-driven visual dubbing methods have achieved great success.
Despite this, we observe that the semantic ambiguity between spatial and temporal domains significantly degrades the synthesis stability for the dynamic faces. 
We argue that aligning the semantic features from spatial and temporal domains is a promising approach to stabilizing facial motion. 
To achieve this, we propose a Spatial-Temporal Semantic Alignment (STSA) method, which introduces a dual-path alignment mechanism and a differentiable semantic representation.
The former leverages a Consistent Information Learning (CIL) module to maximize the mutual information at multiple scales, thereby reducing the manifold differences between spatial and temporal domains. 
The latter utilizes probabilistic heatmap as ambiguity-tolerant guidance to avoid the abnormal dynamics of the synthesized faces caused by slight semantic jittering.
Extensive experimental results demonstrate the superiority of the proposed STSA, especially in terms of image quality and synthesis stability. Pre-trained weights and inference code are available at~\url{https://github.com/SCAILab-USTC/STSA}.

\end{abstract}

\begin{IEEEkeywords}
Audio-driven visual dubbing, Semantic alignment, Multi-modal processing
\end{IEEEkeywords}

\section{Introduction}
Audio-driven visual dubbing has recently garnered attention for applications in digital human creation, video conferencing, and video translation 
\cite{tong2024multimodal, wav2lip, musetalk}.
However, achieving stable motion remains challenging due to the semantic ambiguity between spatial and temporal domains, as the unstructured and motionless nature of driving audio often leads to inconsistency across these domains in the generated semantic regions.

One of the crucial measure to reduce semantic ambiguity is the sufficient utilization of spatial-temporal information. However, most methods perform feature transformations within a single domain, which often leads to insufficiency. Spatial domain-based methods \cite{dinet,adaat} usually stack reference faces along the channel dimension for encoding and then compute affine coefficients for feature deformation of each channel. While this retains more details, it may result in unsmooth face motion. In contrast, temporal domain-based methods \cite{iplap,zhang2021flow, jeong2024seamstalk} typically calculate an optical flow field for each reference face in a cyclic manner, subsequently overlaying the deformed features of each optical flow field. Although optical flow field ensures temporal stability, it is sensitive to image noise and can not handle large displacement.   
Moreover, simply applying spatial and temporal deformations simultaneously can lead to misaligned spatial-temporal features. Since the two domains emphasize different aspects of the same semantic deformation, this misalignment may result in semantic ambiguity, which further disrupts the stability of dynamic facial motion.

On the other hand, semantic representation is also important. Several approaches \cite{wav2lip, dinet, talklip, videoretalking} employ multi-modal feature alignment or fusion strategies to enhance audio-lip movement correlation. However, these methods typically depend on implicit semantic representation, which fails to ensure semantic consistency across consecutive frames.
To further enhance the semantic expression of the synthesized faces, many methods \cite{difftalk, iplap, emmn} introduce semantic-guided generation, which leverages explicit semantic representations to provide structure or motion cues. 
DiffTalk \cite{difftalk} employs keypoints as guidance to provide a coarse structural constraint. 
However, keypoints are insufficient to capture complete semantic variations in the temporal domain.
IP-LAP \cite{iplap} connects keypoints into sketches, increasing the integrality of the semantic structure. 
Nevertheless, the use of pixel-level facial sketches lacks sufficient robustness and smoothness against semantic jitter, often resulting in an inability to smoothly model the facial structure and motion.
Furthermore, since sketch generation and face synthesis are typically trained in two separate stages, the errors from the sketch phase may be accumulated and magnified into the synthesized face video, exacerbating both spatial and temporal stability.

\begin{figure}[t]
    \centering
    \setlength{\abovecaptionskip}{-1mm}
    \includegraphics[width=1\linewidth]{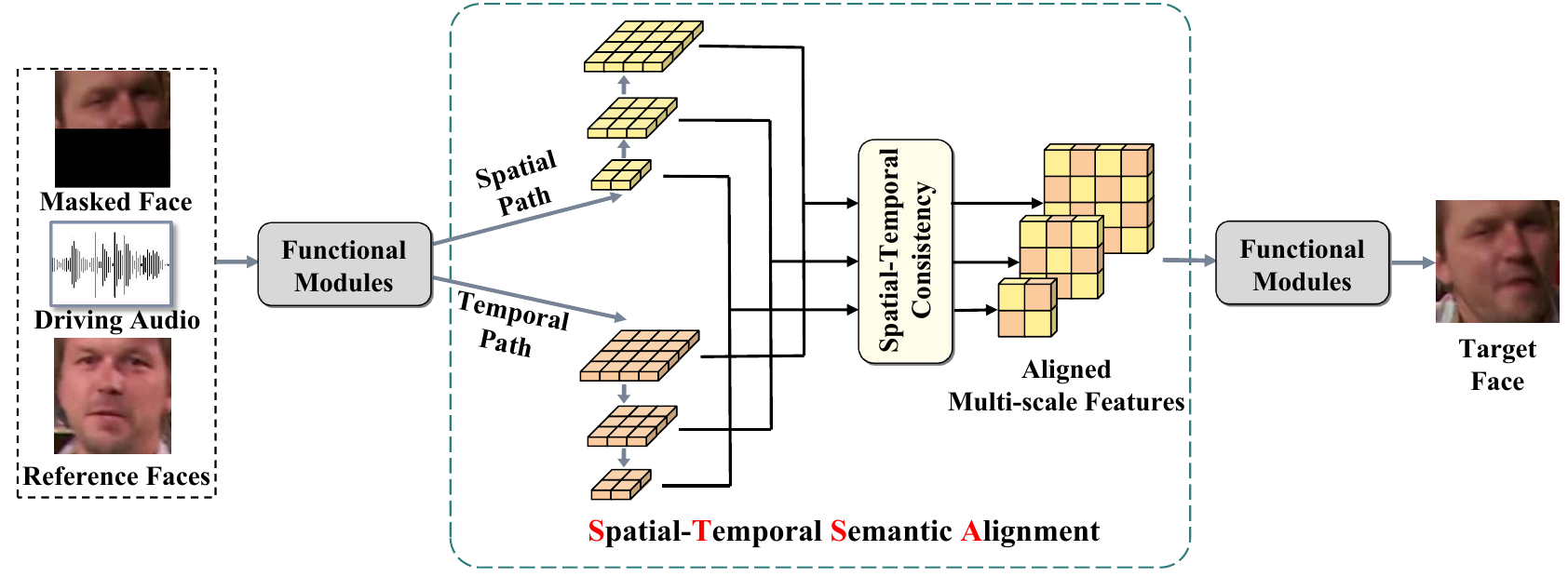}
    \caption{Insight of the proposed STSA. Given the target masked face, driving audio, and reference faces, STSA generates a target face synchronized with audio by aligning the multi-scale features of reference faces in both spatial and temporal domains. \textit{Functional Modules} refer to neural networks used for encoding, decoding, guidance prediction, and face synthesis.}
    \label{fig:f1}
    \vspace{-4mm}
\end{figure}

In this paper, we propose a Spatial-Temporal Semantic Alignment approach, called STSA, to address the challenges mentioned above. The approach promotes semantic consistency between the two domains, enhancing the quality and stability of the synthesized faces, as shown in Figure \ref{fig:f1}. 
In this framework, we design a dual-path alignment mechanism, in which a Consistent Information Learning (CIL) module is introduced to align spatial-temporal information at different scales. CIL can effectively correct misaligned information by maximizing mutual information, enabling us to obtain features that exhibit semantic consistency in both spatial and temporal domains. 
Moreover, we employ the probabilistic heatmap as an ambiguity-tolerant semantic representation to guide the feature deformation of reference faces. It offers malleable guidance, resilient to semantic jittering, ensuring the smoothness of synthesized motion.
Thanks to its differentiability, the guidance prediction and face synthesis stages can be end-to-end secondarily optimized, mitigating the accumulated errors and stage gaps caused by separated optimization.

Our main contributions are summarized as follows:
\begin{itemize}
\item[$\bullet$] We propose a Spatial-Temporal Semantic Alignment (STSA) approach, reducing the semantic ambiguity and achieving realistic and stable visual dubbing.
\item[$\bullet$] We design a dual-path alignment mechanism, which leverages a Consistent Information Learning (CIL) module, maximizing the mutual information across spatial and temporal domains to align them.
\item[$\bullet$] To the best of our knowledge, we are the first to introduce probabilistic heatmap as ambiguity-tolerant semantic guidance, which enhances the smoothness of synthesized motion.
\end{itemize}

\section{Related Work}

\subsection{Semantic-free Visual Dubbing}
Semantic-free approaches \cite{wav2lip, dinet, talklip, videoretalking} without additional guidance primarily achieve high-quality generation through intricate information encoding and decoding, or through specialized model designs.
Wav2Lip \cite{wav2lip} introduces a quality discriminator for adversarial training, and a pre-trained lip-sync expert to supervise the synchronization. 
Wang et al.\cite{talklip} focus on lip movements and propose a lip-reading discriminator to enhance the intelligibility of the generated mouth region.
DINet \cite{dinet} employs dense affine matrices to deform the features of reference faces, thereby obtaining high-quality images.
Due to the absence of explicit semantic guidance, these methods suffer from semantic ambiguity during generation, leading to sub-optimal audio-visual alignment.

\subsection{Semantic-guided Visual Dubbing}
Semantic-guided approaches \cite{difftalk, iplap, emmn, gan2023efficient} can be categorized according to the type of semantic guidance used. 
Approaches \cite{emmn,gan2023efficient} utilize emotion as guidance to facilitate more cohesive alignment between facial expressions and lip movements. EMMN \cite{emmn} extracts emotional features from audio input and retains them for subsequent processing. Such methods necessitate additional emotional labels. 
Some works \cite{difftalk, iplap} use facial keypoints as semantic guidance. DiffTalk \cite{difftalk} employs the diffusion model \cite{ho2020denoising} and utilizes facial keypoints for denoising. 
Techniques \cite{personatalk,geneface} that use 3D information as guidance are mostly based on 3DMM. 
Zhang et al. \cite{personatalk} utilize 3DMM coefficients and attention mechanism to preserve the speaking style.
Ye et al. \cite{geneface} further combine 3DMM with NeRF, achieving improved generation quality. However, NeRF-based methods are person-specific, while this paper focuses on person-general models.

\section{Methodology}
The proposed STSA consists of two stages: guidance prediction and face synthesis, as shown in Figure \ref{fig:f2}. Guidance prediction aims to generate ambiguity-tolerant heatmap from the driving audio. Then, based on the heatmap guidance, face synthesis employs a dual-path alignment mechanism to align multi-scale reference features in both spatial and temporal domains, completing the lower half of the target face.
\subsection{Guidance Prediction}
\begin{figure*}
    \centering
    \setlength{\abovecaptionskip}{-1mm}
    \includegraphics[width=0.9\linewidth]{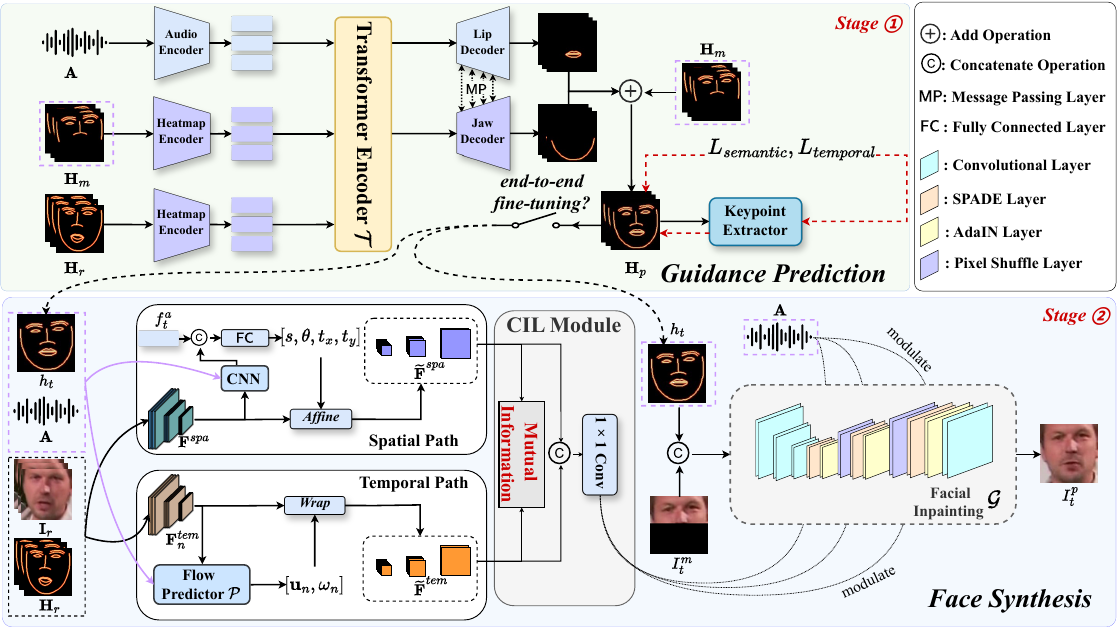}
    \caption{Illustration of the proposed Spatial-Temporal Semantic Alignment (STSA) approach. STSA consists of guidance prediction and face synthesis. The information within the purple dashed box are utilized multiple times throughout the process. The red dashed arrow means supervisions for heatmaps.}
    \label{fig:f2}
    \vspace{-3mm}
\end{figure*}

Given a driving audio and a template video of a target identity, the first step is to generate an audio-aligned heatmap sequence, as shown in the upper half of Figure \ref{fig:f2}. Specifically, we sample audio features $\textbf{A}=\{f^{a}_{t}\}^{T}_{t=1}$, the top-half heatmaps of the target face $\textbf{H}_{m}=\{h^{m}_{t}\}^{T}_{t=1}$, and the reference heatmaps $\textbf{H}_{r}=\{h^{r}_{n}\}^{N}_{n=1}$ to generate the target heatmaps for sequential $T$ frames, where $N$ is the number of reference heatmaps.
$\textbf{H}_{m}$ are used to provide pose information, while $\textbf{H}_{r}$ provide facial structure information. Notably, the two sets of frames selected from the template video are non-overlapping and $\textbf{A}$ are extracted from raw audio using pre-trained Wav2Vec 2.0 \cite{wav2vec}.
Then, we utilize a Transformer encoder \cite{vaswani2017attention} $\mathcal{T}$ to model the temporality and correlations among the inputs, which can be written as follows:
\begin{equation}
\textbf{Z}=\mathcal{T}(\mathcal{H}(\textbf{A}, \textbf{H}_{m}, \textbf{H}_{r}))\label{eq1}
\end{equation}
where $\mathcal{H}(\cdot)$ is the operation that respectively encodes the inputs into tokens using CNNs and concatenates them together, $\textbf{Z}=\{z_{i}\}^{2T+N}_{i=1}$ represent the encoded tokens.

Unlike \cite{iplap}, where a few fully connected layers directly regress $\textbf{Z}$ to keypoint coordinates, we devise a coupled-branch heatmap decoder $\mathcal{D}$ to predict lip and jaw heatmaps. Following \cite{lab}, we introduce Message Passing (MP) layers between corresponding layers of the lip and jaw decoders to pass information, thereby enhancing synchronicity between the lip and jaw regions. The decoding process for each target heatmap is as follows:
\begin{equation}
{h}^{lip}_{t}, {h}^{jaw}_{t}=\mathcal{D}(\mathsf{MP}(z_{t}, z_{t+T}))\label{eq2}
\end{equation}
Then, we can complete the predicted target heatmaps $\textbf{H}_{p}=\{h^{p}_{t}\}^{T}_{t=1}$, where $h^{p}_{t}=h^{m}_{t}+h^{lip}_{t}+h^{jaw}_{t}$.

To supervise the fine-grained semantic structure of the predicted heatmaps, we impose point-level supervision signals to a keypoint set $\textbf{P}_{p}$ that are extracted from the predicted heatmaps using a CNN-based Keypoint Extractor.

Finally, we can align the semantic structures of different granularities by jointly optimizing the prediction of heatmaps and keypoints:
\begin{equation}
\mathcal{L}_{semantic}=\mathbb{E}_{t}(\Vert \textbf{H}_{g} -  \textbf{H}_{p}\Vert_2 + \Vert {\textbf{P}}_{g} -  \textbf{P}_{p}\Vert_2)\label{eq3}
\end{equation}
where $\textbf{H}_{g}$ and $\textbf{P}_{g}$ refer to the ground truths. We also introduce a temporal smoothing loss $\mathcal{L}_{temporal}$ following \cite{iplap}, which constrains the consistency between the adjacent-frame movement of predictions and ground truths. The total loss for this stage is summarized as:
\begin{equation}
\mathcal{L}_{heatmap}=\mathcal{L}_{semantic} + \mathcal{L}_{temporal} \label{eq4}
\end{equation}

\subsection{Face Synthesis}
We design a dual-path alignment mechanism in the face synthesis process to reduce semantic ambiguity between spatial and temporal domains, as shown in the lower half of Figure \ref{fig:f2}.
During the generation of each frame, we utilize the audio feature $f^{a}_{t}$ and the complete heatmap $h_t$ of the target frame to guide the reference feature transformations in both domains. 
Note that during separate training, $h_t$ uses the ground truth, while during end-to-end fine-tuning with guidance prediction and inference, $h_t$ uses the predicted results.

In the spatial path, the reference facial images $\textbf{I}_{r}=\{I^r_n\}^N_{n=1}$ are stacked along the channel dimension and encoded into multi-scale reference features $\textbf{F}^{spa}=\{f^{spa}_{j}\}^{J}_{j=1}$, where $J$, representing the number of scales, is set to $3$. At each scale, we introduce the AdaAT module \cite{adaat} to compute an affine transformation matrix $M_c \in \mathbb{R}^{2\times3}$ for the reference feature of each channel, with four parameters: scale $s$, rotation $\theta$, translation $t_x$ and $t_y$. Here, 
$f^{spa}_{j}$ concatenated with the corresponding heatmaps $\textbf{H}_{r}$ and $h_t$ are sent into a shallow CNN block to obtain pose-aligned features. Following the CNN, a fully connected layer computes the affine transformation matrices for all channels of the features based on its concatenation with $f^{a}_{t}$. This process is illustrated as follows:  
\begin{equation}
\{M_{c}\}^{C}_{c=1} =\mathsf{FC}(f^{a}_{t}\textcircled {c}\mathsf{CNN}(f^{spa}_{j}, \textbf{H}_{r}, h_t)) \label{eq5}
\end{equation}

After this, we transform $\textbf{F}^{spa}$ channel-wise at different scales, obtaining spatial-deformation reference features ${\widetilde{\textbf{F}}^{spa}}=\{\widetilde{f}^{spa}_{j}\}^{J}_{j=1}$.

In the temporal path, inspired by \cite{headgan}, we predict the dense flow from the reference faces to the target face. Dense flow's additivity allows for a comprehensive consideration of motion across multiple frames, enabling more accurate temporal modeling. Distinct from the stacking strategy used in the spatial path, we sequentially concatenate each reference face $I^r_n$ and its corresponding heatmap $h^r_n$, encoding them into multi-scale features $\textbf{F}^{tem}_n=\{f^{tem}_{n,j}\}^{J}_{j=1}$, which are then fed into the flow predictor $\mathcal{P}$. To guide the dense flow transformation to align with the target semantic, we equip AdaIN layers \cite{adain} and SPADE layers \cite{spade} in $\mathcal{P}$ to inject audio feature $f^{a}_{t}$ and target heatmap $h_t$ into the prediction process, respectively:
\begin{equation}
\mathbf{u}_n, \omega_n =\mathcal{P}(\textbf{F}^{tem}_n, \mathsf{AdaIN}(f^{a}_{t}), \mathsf{SPADE}(h_t)) \label{eq6}
\end{equation}
where $\mathbf{u}_n$ and $\omega_n$ denote the dense flow filed and the weight associated with $\textbf{F}^{tem}_n$. Finally, we apply $\mathbf{u}_n$ to wrap $\textbf{F}^{tem}_n$ and compute the weighted sum of the wrapped results to output the temporal-deformation ${\widetilde{\textbf{F}}^{tem}}=\{\widetilde{f}^{tem}_{j}\}^{J}_{j=1}$ of reference features.

\begin{table*}[ht]
\belowrulesep=0pt
\aboverulesep=0pt
\caption{Quantitative results on the LRS2 dataset. $``\uparrow"$ means higher is better while $``\downarrow"$ means lower is better.}
\label{tab:t1}
\centering
{
\begin{tabular}{c|cccc|ccc|c}
\toprule
&\multicolumn{4}{c|}{\textbf{{Visual Quality}}}
&\multicolumn{3}{c|}{\textbf{{Lip-Sync}}}
&\multicolumn{1}{c}{\textbf{{Stablity}}} \\
Method&PSNR$\uparrow$&SSIM$\uparrow$&FID$\downarrow$&LPIPS$\downarrow$&LipLMD$\downarrow$&LSE-D$\downarrow$&LSE-C$\uparrow$&DME$\downarrow$ \\
\midrule
Wav2Lip\cite{wav2lip} &31.30&0.9223&30.67&0.0230&0.00918&\underline{6.4692}&\textbf{8.6729}
&0.8728 \\
PC-AVS\cite{pcavs} &19.13&0.5876&84.86&0.1749&0.03594&6.7751&7.4657&0.8483 \\
DiffTalk\cite{difftalk}&\textbf{32.60}&\textbf{0.9364}&27.15&0.0295&\underline{0.00767}&6.4842&7.7855&0.1452 \\
TalkLip\cite{talklip}&31.22&0.9256&33.49&\underline{0.0209}&0.00923&6.9661&7.2275&0.6896\\
IP-LAP\cite{iplap}&30.59&0.9172&\underline{25.11}&0.0227&0.00883&8.4120&5.1110&0.4187 \\
DINet\cite{dinet} & 25.04 & 0.8003 & 54.14 & 0.0849 & 0.01734 & 7.8405 & 6.0905 & 0.1771 \\
MuseTalk\cite{musetalk} & 30.31 & 0.9058 & 35.21 & 0.0325 & 0.01130 & 7.6472 & 6.4835 & \underline{0.1422} \\
\midrule
Ours & \underline{31.73} & \underline{0.9278} & \textbf{24.23} & \textbf{0.0193} & \textbf{0.00742} & \textbf{6.0970} & \underline{8.3094}&\textbf{0.1346} \\
GT & N/A & 1.0000 & 0.00 & 0.0000 & 0.00000 & 6.3531 & 8.1217 & 0.0000     \\
\bottomrule
\end{tabular}
\vspace{-3mm}
}
\end{table*}
\begin{figure*}[ht]
    \centering
    \setlength{\abovecaptionskip}{-2mm}
    \includegraphics[width=0.85\linewidth]{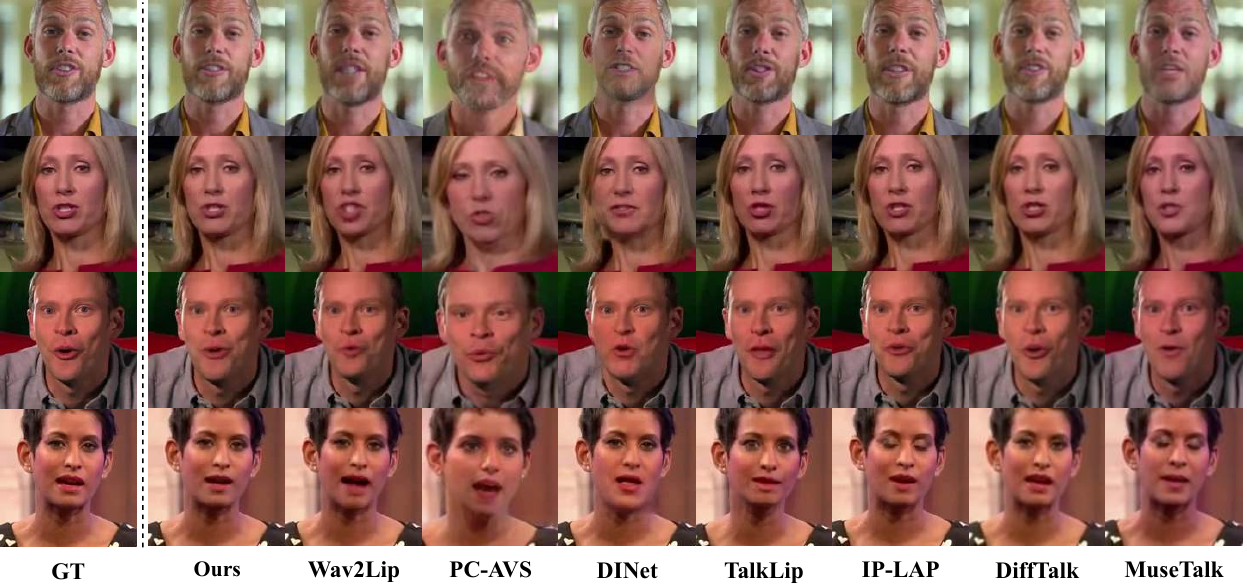}
    \caption{Qualitative results on LRS2. Please zoom in to get more details.}
    \label{fig:f3}
    \vspace{-6mm}
\end{figure*}
After obtaining the two sets of reference features, we leverage a Consistency Information Learning (CIL) module to align them by maximizing their mutual information. In detail, we encode the spatial and temporal features into embeddings $E^{spa}$ and $E^{tem}$, then maximizing the lower-bound based on the Donsker-Varadhan representation \cite{donsker} of the KL-divergence by the following loss function as \cite{mine,dim}:
\begin{equation}
\mathcal{L}_{mi} = -(\mathbb{E}_{\mathbb{J}}[\textbf{T}_{\theta}(E^{spa}, E^{tem})] - \mathrm{log}\mathbb{E}_{\mathbb{M}}[e^{\textbf{T}_{\theta}(E^{spa}, E^{tem})}])\label{eq7}
\end{equation}
where $\mathbb{E}_{\mathbb{J}}$ and $\mathbb{E}_{\mathbb{M}}$ are the expectations of the joint distribution and the marginal distribution, respectively. $\textbf{T}_{\theta}$ is a learnable neural network. Then, ${\widetilde{\textbf{F}}^{spa}}$ and ${\widetilde{\textbf{F}}^{tem}}$ are fused through a $1\times1$ convolutional layer to obtain aligned reference features $\textbf{F}^{align}$. Please refer to the supplementary materials for more details.

We then construct a facial inpainting module $\mathcal{G}$ following Face-HeadGAN \cite{headgan} that accepts a target masked face $I^{m}_t$ and target heatmap $h_t$ as inputs to render the final face $I^{p}_t$. During this process, the audio feature and the aligned reference features are infused into the rendering process using AdaIN and SPADE layers, which provide comprehensive structural and textural information for the masked areas:
\begin{equation}
I^{p}_t =\mathcal{G}(I^{m}_t, h_t, \mathsf{AdaIN}(f^{a}_{t}), \mathsf{SPADE}(\textbf{F}^{align})) \label{eq8}
\end{equation}

To enhance the realism of generated face, we incorporate perceptual loss:
\begin{equation}
\mathcal{L}_{perc}=\sum_i\Vert \phi_i(I^{p}_t)-\phi_i(I^{g}_t) \Vert_1 +\Vert G^{\phi}_i(I^{p}_t)-G^{\phi}_i(I^{g}_t) \Vert_1 \label{eq9}
\end{equation}
where $\phi_i$ is the $i$-th layer of the pre-trained VGG-19 \cite{vgg} and $ G^{\phi}_i$ is the Gram matrix of the $\phi_i$'s output, which is used to constrain the style consistency between the generated face $I^{p}_t$ and the ground truth $I^{g}_t$.

To further improve the image details, we introduce patch GAN loss $\mathcal{L}_{gan}$ and feature matching loss $\mathcal{L}_{feat}$ following \cite{iplap} and \cite{headgan}. Moreover, a pre-trained SyncNet is utilized to compute sync loss $\mathcal{L}_{sync}$ as \cite{wav2lip}, which constrains the audio-lip synchronization. Finally, we sum up all losses to define the total loss function:
\begin{equation}
\mathcal{L}_{face}=\lambda_1\mathcal{L}_{perc}+\lambda_2\mathcal{L}_{gan}+\lambda_3\mathcal{L}_{feat}+\lambda_4\mathcal{L}_{sync}+\lambda_5\mathcal{L}_{mi} \label{eq10}
\end{equation}

Due to the learnable heatmap prediction, we can further secondarily optimize the whole model end-to-end, which significantly reduces the accumulation of errors from the guidance prediction into the face synthesis.

\section{Experiments}
\subsection{Datasets and Implementation Details}
\textbf{Datasets.}
In the experiments, we select two widely used datasets, LRS2 \cite{lrs2} and CMLR \cite{cmlr}. LRS2 is a large English dataset with various head poses in real-word settings, containing news and talk shows from BBC programs. CMLR, collected from nation news program, consists of numerous Chinese Mandarin spoken sentences with frontal head poses. In this setting, only the training set of LRS2 is used to train the models, while testing data are randomly selected from the test sets of LRS2 and CMLR. Here, 40 test videos from LRS2 are evaluated for intra-domain testing, and 40 test videos from CMLR for cross-domain testing.

\textbf{Implementation Details.}
In the data pre-processing stage, all the videos are converted to 25 FPS and the audios are resampled to 16KHz. We utilize mediapipe \cite{mediapipe} to detect faces and landmarks. Then the faces are resized to $128\times128$ and the landmarks are transformed to heatmap as \cite{lab}. During trainng, the number of reference input $N$ is set to 5. The weights of the loss term in face synthesis are set to: $\lambda_1=4, \lambda_2=0.25, \lambda_3=1, \lambda_4=0.1, \lambda_5=0.01.$
For more implementation details, please refer to the supplementary materials.

\textbf{Evaluation Metrics.}
We choose the metrics of Peak Signal-to-Noise Ratio (PSNR), Structural Similarity (SSIM) \cite{ssim}, Learned Perceptual Image Patch Similarity (LPIPS) \cite{lpips} to evaluate the visual quality. Besides, we compute Fr${\rm \acute{e}}$chet Inception Distance (FID) \cite{fid} to evaluate the realism of the synthesized results. LSE-C, LSE-D \cite{wav2lip} and the normalized lip landmarks distance (LipLMD) are adopted to evaluate the audio-lip synchronization. For the motion stability, we follow Dynamic Motion Error (DME)\cite{dyco} as the metric. 

\begin{table}[t]
\vspace{-2.0mm}
\belowrulesep=0pt
\aboverulesep=0pt
\caption{Quantitative results on the CMLR dataset.}
\label{tab:t2}
\centering
\setlength{\tabcolsep}{3.2pt} 
\resizebox{1\linewidth}{!}{
\begin{tabular}{c|cccc|c|c}
\toprule
&\multicolumn{4}{c|}{\textbf{{Visual Quality}}}
&\multicolumn{1}{c|}{\textbf{{Lip-Sync}}}
&\multicolumn{1}{c}{\textbf{{Stablity}}} \\
Method&PSNR$\uparrow$&SSIM$\uparrow$&FID$\downarrow$&LPIPS$\downarrow$&LipLMD$\downarrow$&DME$\downarrow$ \\
\midrule
Wav2Lip\cite{wav2lip}&29.22&0.9223&24.04&0.0176&0.00838&0.0515  \\
PC-AVS\cite{pcavs}&16.30&0.4826&143.58&0.2863&0.09561&0.4061 \\
DiffTalk\cite{difftalk}&25.58&0.8940&23.17&0.0374&0.00821&0.0315\\
TalkLip\cite{talklip}&29.18&\textbf{0.9314}&26.87&\underline{0.0150}&0.00769&0.0598\\
IP-LAP\cite{iplap}&\underline{29.38}&0.9215&\underline{17.67}&0.0156&\underline{0.00739}&0.0517 \\
DINet\cite{dinet} &26.65&0.8703&18.77&0.0331&0.00994&\textbf{0.0219}   \\
MuseTalk\cite{musetalk} & 28.13 & 0.8965 & 24.51 & 0.0219 & 0.01099 & 0.0271 \\
\midrule
Ours &\textbf{30.21}&\underline{0.9280}&\textbf{17.07}&\textbf{0.0140}&\textbf{0.00656}&\underline{0.0255} \\
GT & N/A &1.0000&0.00&0.0000&0.00000&0.0000   \\
\bottomrule
\end{tabular} }
\vspace{-3mm}
\end{table}

\subsection{Comparisons with State-of-the-Art Works}
We choose some state-of-the-art methods for comparison: Wav2Lip \cite{wav2lip}, PC-AVS\cite{pcavs}, DiffTalk\cite{difftalk}, TalkLip\cite{talklip}, IP-LAP \cite{iplap}, DINet \cite{dinet} and MuseTalk \cite{musetalk}. The quantitative results on LRS2 and CMLR are presented in Table \ref{tab:t1} and \ref{tab:t2}. The qualitative results are shown in Figure \ref{fig:f3}.

From Table \ref{tab:t1}, it can be seen that our method demonstrates superiority over other methods, particularly in terms of image quality and stability. The LSE-C of Wav2Lip is slightly higher than ours, but we are closer to the ground truth, and our image quality is superior to theirs.
The results in Table \ref{tab:t2} show that STSA exhibits excellent generalization ability, which indicates that STSA effectively aligns the spatial and temporal domains and reduces semantic ambiguity. 

Although Difftalk outperforms us in PSNR and SSIM on the test set from the same domain as the training set, it may have overfitted, resulting in a sharp decline in performance for cross-dataset evaluation. Moreover, its realism (FID) is inferior to ours. 
DINet achieves higher stability on the CMLR, which mainly consists of frontal head poses, but its performance significantly drops on the LRS2 with various head poses due to the limitations of its alignment mechanism.

Figure \ref{fig:f3} presents a qualitative comparison of the results generated by our method and other approaches. Visually, the results generated by our method are closest to the ground truth for both image quality and lip-sync accuracy.

\begin{table}[t]
\vspace{-2.0mm}
\belowrulesep=0pt
\aboverulesep=0pt
\caption{Ablation study. FT denotes the end-to-end fine-tuning for guidance prediction and face synthesis.}
\label{tab:t3}
\centering

\resizebox{\linewidth}{!}
{
\begin{tabular}{c|ccccc}
\toprule
&PSNR$\uparrow$&SSIM$\uparrow$&FID$\downarrow$&LSE-C$\uparrow$&DME$\downarrow$ \\
\midrule
w/o CIL &31.22&0.9243&24.63&8.0476&0.1377\\
w/o sync loss &31.11&0.9237&25.32&7.6470&0.1376\\
w/o spatial &30.66&0.9196&25.37&7.1514&0.1371\\
w/o temporal &29.12&0.9036&30.83&8.0844&0.1405 \\
\midrule
Ours (w/o FT) &31.25&0.9253&24.87&8.1446&0.1362  \\
Ours (w/ FT) &\textbf{31.73}&\textbf{0.9278}&\textbf{24.23}&\textbf{8.3094}&\textbf{0.1346}\\
\bottomrule
\end{tabular} 
}
\vspace{-3mm}
\end{table}

\subsection{Ablation Study.}
To investigate the key components in our model, we conduct an ablation study. Specifically, we design six sets of experiments, including: (1) without CIL (w/o CIL); (2) without sync loss (w/o sync loss); (3) without spatial (w/o spatial), where the spatial deformation path is removed; (4) without temporal (w/o temporal), where the temporal deformation path is removed; (5) without fine-tuning (w/o FT), where guidance prediction and face synthesis are trained separately; (6) with fine-tuning (w/ FT), where end-to-end fine-tuning is performed. Note that (1)–(4) are also not fine-tuned.

\begin{figure}[t]
    \centering
    \setlength{\abovecaptionskip}{-2mm}
    \includegraphics[width=0.8\linewidth]{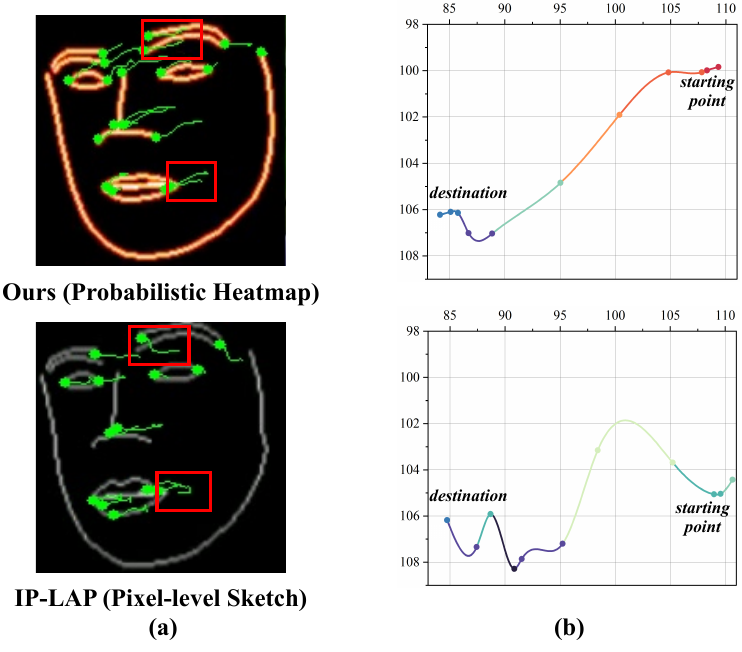}
    \caption{We visualize the motion trajectories of the heatmap in this paper and the sketch in IP-LAP with the Lucas Kanade algorithm, as shown in (a). Furthermore, we plot the trajectories of the right mouth corner in the coordinate system in (b).}
    \label{fig:f4}
    \vspace{-6mm}
\end{figure}

From Table \ref{tab:t3}, we observe the following: Removing the CIL module reduces overall performance, demonstrating the effectiveness of CIL in aligning spatial-temporal domains. The supervision of pre-trained SyncNet helps to improve synchronization. The spatial deformation path contributes to synchronization improvement, while the temporal deformation path enhances image quality and stability, suggesting that they play a complementary role. 
Key performance indicators are overall improved after two-stage end-to-end fine-tuning, indicating that this fine-tuning eliminates error accumulation across stages and further reduces semantic ambiguity. More ablation results are represented in the supplementary materials.

\subsection{Semantic Guidance Comparison}
In terms of intermediate semantic guidance, our probabilistic heatmap is compared with the pixel-level sketch in IP-LAP. We track the corner points of semantic guidance inferred by both methods over the same motion sequence using the Lucas-Kanade algorithm \cite{LK}. As shown in Figure \ref{fig:f4}, the motion trajectory from our heatmap is smoother. In contrast, the trajectory from the sketch is more convoluted, which may lead to apparent semantic ambiguity.
Note that the differentiable heatmap enables end-to-end fine-tuning for the whole model, which is impossible with existing sketch guidance.
\subsection{Limitations} 
There is a rhythmic coupling between human head movements and speech. When we select cross-domain audio to drive the template video, the rhythmic misalignment caused by the domain shift from the audio may lead to noticeable artifacts. Additionally, our inference efficiency is slightly lower than the single-path methods.

\section{Conclusion}
In this paper, we have proposed a Spatial-Temporal Semantic Alignment approach, named STSA, which significantly improves the generation quality and motion stability of visual dubbing by reducing semantic ambiguity. Specifically, STSA introduces a dual-path alignment mechanism and a differentiable semantic representation, both of which contribute to improving semantic ambiguity issues. Experimental results demonstrate that STSA achieves competitive performance compared to state-of-the-art methods. Future work will focus on improving the efficiency of generation.

\bibliographystyle{IEEEtran}
\bibliography{IEEEabrv,ICME_citation}

\begin{thebibliography}{10}
\providecommand{\url}[1]{#1}
\csname url@samestyle\endcsname
\providecommand{\newblock}{\relax}
\providecommand{\bibinfo}[2]{#2}
\providecommand{\BIBentrySTDinterwordspacing}{\spaceskip=0pt\relax}
\providecommand{\BIBentryALTinterwordstretchfactor}{4}
\providecommand{\BIBentryALTinterwordspacing}{\spaceskip=\fontdimen2\font plus
\BIBentryALTinterwordstretchfactor\fontdimen3\font minus \fontdimen4\font\relax}
\providecommand{\BIBforeignlanguage}[2]{{%
\expandafter\ifx\csname l@#1\endcsname\relax
\typeout{** WARNING: IEEEtran.bst: No hyphenation pattern has been}%
\typeout{** loaded for the language `#1'. Using the pattern for}%
\typeout{** the default language instead.}%
\else
\language=\csname l@#1\endcsname
\fi
#2}}
\providecommand{\BIBdecl}{\relax}
\BIBdecl

\bibitem{tong2024multimodal}
H.~Tong, H.~Li, H.~Du, Z.~Yang, C.~Yin, and D.~Niyato, ``Multimodal semantic communication for generative audio-driven video conferencing,'' \emph{IEEE Wireless Communications Letters}, pp. 1--1, 2024.

\bibitem{wav2lip}
K.~Prajwal, R.~Mukhopadhyay, V.~P. Namboodiri, and C.~Jawahar, ``A lip sync expert is all you need for speech to lip generation in the wild,'' in \emph{MM}.\hskip 1em plus 0.5em minus 0.4em\relax ACM, 2020, pp. 484--492.

\bibitem{musetalk}
Y.~Zhang, M.~Liu, Z.~Chen, B.~Wu, Y.~Zeng, C.~Zhan, Y.~He, J.~Huang, and W.~Zhou, ``Musetalk: Real-time high quality lip synchronization with latent space inpainting,'' \emph{ArXiv}, vol. abs/2410.10122, 2024.

\bibitem{dinet}
Z.~Zhang, Z.~Hu, W.~Deng, C.~Fan, T.~Lv, and Y.~Ding, ``Dinet: Deformation inpainting network for realistic face visually dubbing on high resolution video,'' in \emph{AAAI}, vol.~37, no.~3, 2023, pp. 3543--3551.

\bibitem{adaat}
Z.~Zhang and Y.~Ding, ``Adaptive affine transformation: A simple and effective operation for spatial misaligned image generation,'' in \emph{MM}.\hskip 1em plus 0.5em minus 0.4em\relax ACM, 2022, pp. 1167--1176.

\bibitem{iplap}
W.~Zhong, C.~Fang, Y.~Cai, P.~Wei, G.~Zhao, L.~Lin, and G.~Li, ``Identity-preserving talking face generation with landmark and appearance priors,'' in \emph{CVPR}.\hskip 1em plus 0.5em minus 0.4em\relax IEEE, 2023, pp. 9729--9738.

\bibitem{zhang2021flow}
Z.~Zhang, L.~Li, Y.~Ding, and C.~Fan, ``Flow-guided one-shot talking face generation with a high-resolution audio-visual dataset,'' in \emph{CVPR}.\hskip 1em plus 0.5em minus 0.4em\relax IEEE, 2021, pp. 3661--3670.

\bibitem{jeong2024seamstalk}
Y.~Jeong, G.~Kim, D.~Jang, J.~Hwang, and E.~Yang, ``Seamstalk: Seamless talking face generation via flow-guided inpainting,'' \emph{IEEE Access}, 2024.

\bibitem{talklip}
J.~Wang, X.~Qian, M.~Zhang, R.~T. Tan, and H.~Li, ``Seeing what you said: Talking face generation guided by a lip reading expert,'' in \emph{CVPR}.\hskip 1em plus 0.5em minus 0.4em\relax IEEE, 2023, pp. 14\,653--14\,662.

\bibitem{videoretalking}
K.~Cheng, X.~Cun, Y.~Zhang, M.~Xia, F.~Yin, M.~Zhu, X.~Wang, J.~Wang, and N.~Wang, ``Videoretalking: Audio-based lip synchronization for talking head video editing in the wild,'' in \emph{SIGGRAPH Asia}.\hskip 1em plus 0.5em minus 0.4em\relax ACM, 2022, pp. 1--9.

\bibitem{difftalk}
S.~Shen, W.~Zhao, Z.~Meng, W.~Li, Z.~Zhu, J.~Zhou, and J.~Lu, ``Difftalk: Crafting diffusion models for generalized audio-driven portraits animation,'' in \emph{CVPR}.\hskip 1em plus 0.5em minus 0.4em\relax IEEE, 2023, pp. 1982--1991.

\bibitem{emmn}
S.~Tan, B.~Ji, and Y.~Pan, ``Emmn: Emotional motion memory network for audio-driven emotional talking face generation,'' in \emph{ICCV}.\hskip 1em plus 0.5em minus 0.4em\relax IEEE, 2023, pp. 22\,089--22\,099.

\bibitem{gan2023efficient}
Y.~Gan, Z.~Yang, X.~Yue, L.~Sun, and Y.~Yang, ``Efficient emotional adaptation for audio-driven talking-head generation,'' in \emph{CVPR}.\hskip 1em plus 0.5em minus 0.4em\relax IEEE, 2023, pp. 22\,634--22\,645.

\bibitem{ho2020denoising}
J.~Ho, A.~Jain, and P.~Abbeel, ``Denoising diffusion probabilistic models,'' \emph{NeurIPS}, vol.~33, pp. 6840--6851, 2020.

\bibitem{personatalk}
L.~Zhang, S.~Liang, Z.~Ge, and T.~Hu, ``Personatalk: Bring attention to your persona in visual dubbing,'' \emph{arXiv}, 2024.

\bibitem{geneface}
Z.~Ye, Z.~Jiang, Y.~Ren, J.~Liu, J.~He, and Z.~Zhao, ``Geneface: Generalized and high-fidelity audio-driven 3d talking face synthesis,'' \emph{arXiv}, 2023.

\bibitem{wav2vec}
A.~Baevski, Y.~Zhou, A.~Mohamed, and M.~Auli, ``wav2vec 2.0: A framework for self-supervised learning of speech representations,'' \emph{NeurIPS}, vol.~33, pp. 12\,449--12\,460, 2020.

\bibitem{vaswani2017attention}
A.~Vaswani, N.~M. Shazeer, N.~Parmar, J.~Uszkoreit, L.~Jones, A.~N. Gomez, L.~Kaiser, and I.~Polosukhin, ``Attention is all you need,'' \emph{NeurIPS}, 2017.

\bibitem{lab}
W.~Wu, C.~Qian, S.~Yang, Q.~Wang, Y.~Cai, and Q.~Zhou, ``Look at boundary: A boundary-aware face alignment algorithm,'' in \emph{CVPR}.\hskip 1em plus 0.5em minus 0.4em\relax IEEE, 2018, pp. 2129--2138.

\bibitem{headgan}
M.~C. Doukas, E.~Ververas, V.~Sharmanska, and S.~Zafeiriou, ``Free-headgan: Neural talking head synthesis with explicit gaze control,'' \emph{TPAMI}, vol.~45, no.~8, pp. 9743--9756, 2023.

\bibitem{adain}
X.~Huang and S.~Belongie, ``Arbitrary style transfer in real-time with adaptive instance normalization,'' in \emph{CVPR}.\hskip 1em plus 0.5em minus 0.4em\relax IEEE, 2017, pp. 1501--1510.

\bibitem{spade}
T.~Park, M.-Y. Liu, T.-C. Wang, and J.-Y. Zhu, ``Semantic image synthesis with spatially-adaptive normalization,'' in \emph{CVPR}.\hskip 1em plus 0.5em minus 0.4em\relax IEEE, 2019, pp. 2337--2346.

\bibitem{pcavs}
H.~Zhou, Y.~Sun, W.~Wu, C.~C. Loy, X.~Wang, and Z.~Liu, ``Pose-controllable talking face generation by implicitly modularized audio-visual representation,'' in \emph{CVPR}.\hskip 1em plus 0.5em minus 0.4em\relax IEEE, 2021, pp. 4176--4186.

\bibitem{donsker}
M.~D. Donsker and S.~S. Varadhan, ``Asymptotic evaluation of certain markov process expectations for large time. iv,'' \emph{Communications on pure and applied mathematics}, vol.~36, no.~2, pp. 183--212, 1983.

\bibitem{mine}
M.~I. Belghazi, A.~Baratin, S.~Rajeswar, S.~Ozair, Y.~Bengio, A.~Courville, and R.~D. Hjelm, ``Mine: mutual information neural estimation,'' \emph{arXiv}, 2018.

\bibitem{dim}
R.~D. Hjelm, A.~Fedorov, S.~Lavoie-Marchildon, K.~Grewal, P.~Bachman, A.~Trischler, and Y.~Bengio, ``Learning deep representations by mutual information estimation and maximization,'' in \emph{ICLR}, 2019.

\bibitem{vgg}
K.~Simonyan, ``Very deep convolutional networks for large-scale image recognition,'' \emph{arXiv}, 2014.

\bibitem{lrs2}
T.~Afouras, J.~S. Chung, A.~Senior, O.~Vinyals, and A.~Zisserman, ``Deep audio-visual speech recognition,'' \emph{TPAMI}, vol.~44, no.~12, pp. 8717--8727, 2018.

\bibitem{cmlr}
Y.~Zhao, R.~Xu, and M.~Song, ``A cascade sequence-to-sequence model for chinese mandarin lip reading,'' in \emph{MM Asia}.\hskip 1em plus 0.5em minus 0.4em\relax ACM, 2019, pp. 1--6.

\bibitem{mediapipe}
C.~Lugaresi, J.~Tang, H.~Nash, C.~McClanahan, E.~Uboweja, M.~Hays, F.~Zhang, C.-L. Chang, M.~G. Yong, J.~Lee, W.-T. Chang, W.~Hua, M.~Georg, and M.~Grundmann, ``Mediapipe: A framework for building perception pipelines,'' \emph{ArXiv}, vol. abs/1906.08172, 2019.

\bibitem{ssim}
Z.~Wang, A.~C. Bovik, H.~R. Sheikh, and E.~P. Simoncelli, ``Image quality assessment: from error visibility to structural similarity,'' \emph{TIP}, vol.~13, no.~4, pp. 600--612, 2004.

\bibitem{lpips}
R.~Zhang, P.~Isola, A.~A. Efros, E.~Shechtman, and O.~Wang, ``The unreasonable effectiveness of deep features as a perceptual metric,'' in \emph{CVPR}.\hskip 1em plus 0.5em minus 0.4em\relax IEEE, 2018, pp. 586--595.

\bibitem{fid}
M.~Heusel, H.~Ramsauer, T.~Unterthiner, B.~Nessler, and S.~Hochreiter, ``Gans trained by a two time-scale update rule converge to a local nash equilibrium,'' \emph{NeurIPS}, vol.~30, 2017.

\bibitem{dyco}
Y.~Chen, Y.~Zhan, Z.~Zhong, W.~Wang, X.~Sun, Y.~Qiao, and Y.~Zheng, ``Within the dynamic context: Inertia-aware 3d human modeling with pose sequence,'' in \emph{ECCV}.\hskip 1em plus 0.5em minus 0.4em\relax Springer, 2025, pp. 491--508.

\bibitem{LK}
B.~D. Lucas and T.~Kanade, ``An iterative image registration technique with an application to stereo vision,'' in \emph{IJCAI}, 1981.

\end{thebibliography}

\end{document}